\documentclass[pmlr]{jmlr} 
    
  \usepackage{longtable} 
\usepackage{wrapfig}

      \usepackage{booktabs}
      \usepackage[load-configurations=version-1]{siunitx}    \usepackage{cleveref}

  \theorembodyfont{\upshape}
\theoremheaderfont{\scshape}
\theorempostheader{:}
\theoremsep{\newline}

  \jmlrvolume{}
\jmlryear{}
\jmlrworkshop{Machine Learning Competitions}

\title[MineRL 2020: robust and domain agnostic reinforcement learning]{Towards robust and domain agnostic\\  reinforcement learning competitions: MineRL 2020}

  \author{
\Name{William Hebgen Guss} \Email{wguss@\{openai.com,cs.cmu.edu\}}\\
\Name{Stephanie Milani} \Email{smilani@cs.cmu.edu}\\
\Name{Nicholay Topin} \Email{ntopin@cs.cmu.edu}\\
\Name{Brandon Houghton} \Email{brandon@openai.com}\\
\Name{Sharada Mohanty} \Email{mohanty@aicrowd.com}\\\\
\Name{Andrew Melnik} \Email{andrew.melnik.papers@gmail.com} \\
\Name{Augustin Harter} \Email{aharter@techfak.uni-bielefeld.de} \\
\Name{Benoit Buschmaas} \Email{bbuschmaas@techfak.uni-bielefeld.de}\\
\Name{Bjarne Jaster} \Email{bjaster@techfak.uni-bielefeld.de}\\
\Name{Christoph Berganski} \Email{cberganski@techfak.uni-bielefeld.de}\\
\Name{Dennis Heitkamp} \Email{dheitkamp@techfak.uni-bielefeld.de}\\
\Name{Marko Henning} \Email{mhenning@techfak.uni-bielefeld.de}\\
\Name{Helge Ritter} \Email{helge@techfak.uni-bielefeld.de} \\
\Name{Chengjie Wu} \Email{wucj19@mails.tsinghua.edu.cn}\\
\Name{Xiaotian Hao} \Email{xiaotianhao@tju.edu.cn}\\
\Name{Yiming Lu} \Email{luym19@mails.tsinghua.edu.cn}\\
\Name{Hangyu Mao} \Email{maohangyu1@huawei.com}\\
\Name{Yihuan Mao} \Email{maoyh20@mails.tsinghua.edu.cn}\\
\Name{Chao Wang} \Email{wangchao358@huawei.com}\\
\Name{Michal Opanowicz} \Email{michal.opanowicz@gmail.com}\\
\Name{Anssi Kanervisto} \Email{anssk@uef.fi}\\
\Name{Yanick Schraner} \Email{yanick.schraner@fhnw.ch}\\
\Name{Christian Scheller} \Email{christian.scheller@fhnw.ch}\\
\Name{Xiren Zhou} \Email{xz2754@columbia.edu}\\
\Name{Lu Liu} \Email{liu@platoapp.com}\\
\Name{Daichi Nishio} \Email{dnish240@gmail.com}\\
\Name{Toi Tsuneda} \Email{ttsuneda@csl.ec.t.kanazawa-u.ac.jp}\\
\Name{Karolis Ramanauskas} \Email{karolis.ram@gmail.com}\\
\Name{Gabija Juceviciute} \Email{g.juceviciute@gmail.com }
  }

    \newcommand{\rlil}{RL + IL }
   \newcommand{\ilonly}{IL-only }
\newcommand{\team}[1]{{{\texttt{#1}}}}

\editor{}

\begin{document}

\maketitle

\begin{abstract}
   Reinforcement learning competitions have formed the basis for standard research benchmarks, galvanized advances in the state-of-the-art, and shaped the direction of the field. Despite this, a majority of challenges suffer from the same fundamental problems: participant solutions to the posed challenge are usually domain-specific, biased to maximally exploit compute resources, and not guaranteed to be reproducible. In this paper, we present a new framework of competition design that promotes the development of algorithms that overcome these barriers.  We propose four central mechanisms for achieving this end: submission retraining, domain randomization, desemantization through domain obfuscation, and the limitation of competition compute and environment-sample budget. To demonstrate the efficacy of this design, we proposed, organized, and ran the MineRL 2020 Competition on Sample-Efficient Reinforcement Learning. In this work, we describe the organizational outcomes of the competition and show that the resulting participant submissions are reproducible, non-specific to the competition environment, and sample/resource efficient, despite the difficult competition task.
\end{abstract}

\section{Introduction}
\label{sec:intro}

 Deep reinforcement learning has emerged as a compelling solution to a wide range of problems in machine learning.
    Techniques from this field have been successfully applied to a number of difficult domains such as real-time video games~\citep{berner2019dota, vinyals2019grandmaster}, complicated control and scheduling problems, real-world robotic manipulation tasks, and self-driving.
The success of deep reinforcement learning (RL) has been accompanied by an increase in RL competitions spanning a number of domains and difficult open problems~\citep{gussminerlneurips2019, perez2019multi, koppejan2009neuroevolutionary}. 
    As competitions mature, they form the basis for  benchmarks used throughout the community~\citep{machado2018revisiting, bellemare2013arcade}.
Typically, competitions focus on a core problem with current RL algorithms or domain(s) not yet readily solved by current methods, and challenge competitors to train agents (using competitor resources) to solve domain(s). These agents are then submitted to an evaluation platform and ranked based on final performance.

 Despite this common format for research benchmarks, a number of issues have become apparent.
    In competitions where only final trained agents are submitted, the algorithmic underpinnings of submissions become difficult to reproduce: 
        winning solutions are often trained with a disproportionately larger compute resource budget to that of other competitors~\citep{gussminerlneurips2019}; 
        competitors often choose to train on a specific set of environment seeds and benefit greatly from large-scale hyperparameter searches, making reimplementation and broader use more difficult~\citep{khetarpal2018re};
        and training code is sometimes not shared when only inference code is submitted, preventing validation of the algorithmic claims of the submission and allowing the submission to be trained using hard-coded, engineered features or action and reward shaping~\citep{houghton2020guaranteeing}. 
     Furthermore, competitions tied to a specific unrandomized domain can fail to yield direct algorithmic advancements, as the most successful methods commonly overfit to and exploit the specific structure of the problem. 
    Similarly, multi-year competitions reward increases in domain knowledge exploitation, reducing the role of algorithmic novelty. 
    For domains of specific real-world importance, such as robotics or self-driving, the task solution has greater utility than any resulting secondary algorithmic advancements. 
    However, there is a large gap between domain-specific submissions to RL competitions on video-game or artificial domains and their downstream utility in the research community.

To address these problems, we proposed, organized, and ran the MineRL 2020 Competition on Sample Efficient Reinforcement Learning using Human Priors~\citep{guss2021minerl}. 
Our competition utilized several novel mechanisms for yielding robust and domain agnostic submissions, including observation and action space obfuscation, submission retraining, domain randomization, and environment interaction limits. 
In this paper, we present the general methodologies and design principles that comprise the competition structure and describe the resulting top-performing submissions. 
\Cref{sec:background} provides a general background for the problem settings that motivate the competition. 
In~\Cref{sec:competition}, we give an overview of the competition, including its design, central task, rules, and resources provided.
In~\Cref{sec:solutions}, the top teams describe the approaches used in their submissions. 
Thereafter, in~\Cref{sec:discussion} we discuss the organizational outcomes of our competition with respect to our goal of robust, reproducible, and high quality solutions. Finally, we position the MineRL competition in the context of other concurrent and past RL competitions in~\Cref{sec:related} and discuss challenges and opportunities for future work in~\Cref{sec:conclusion}.

\section{Background}
\label{sec:background}

\subsection{The sample \emph{in}efficiency problem}
Many of the most celebrated successes of machine learning, such as AlphaStar~\citep{starcraft2019}, AlphaGo~\citep{silver2017mastering}, OpenAI Five~\citep{berner2019dota}, and their derivative systems~\citep{alphazero}, utilize deep RL to achieve human or super-human level performance in sequential decision-making tasks.
These improvements to the state-of-the-art have thus far required exponentially increasing computational power~\citep{amodei_hednandez_2018}, which is  largely due to the number of environment-samples required for training. 
 These growing computational requirements prohibit many in the AI community from  improving these systems and reproducing state-of-the-art results. 
Additionally, the application of many reinforcement learning techniques to real-world challenges, such as self-driving vehicles, is hindered by the raw number of required samples.

A variety of approaches have been proposed towards the goal of sample efficiency, including learning a model of the environment~\citep{buckman2018sample}, leveraging AutoRL to perform efficient hyperparameter optimization~\citep{franke2020sample}, and incorporating domain information in the form of human priors and demonstrations~\citep{dubey2018investigating,pfeiffer2018reinforced}. 
In our competition we encourage the use of any techniques that improve the sample efficiency of RL algorithms without using domain-specific hard-coding.  
 
\subsection{Generalization}
The development of RL algorithms that can generalize is of great importance to the research community~\citep{zhang2018dissection,cobbe2019quantifying,malik2021generalizable}.
There are various notions of generalization, but it is broadly defined as the ability of an agent to learn desired behavior and perform well in similar environments. 
 Through our competition, we want to promote the development of algorithms that generalize across different domains and tasks, such as those with different state and action spaces.

\subsection{Minecraft}
The central competition task, \texttt{ObtainDiamond}, is set in the Minecraft domain. 
 Minecraft is a popular and compelling environment for the development of reinforcement~\citep{oh2016control,shu2017hierarchical,tessler2017deep} and imitation learning methods because of the unique challenges it presents.
 Notably, the procedurally-generated world is composed of discrete blocks that allow modification.
Over the course of gameplay, players change their surroundings by gathering resources and constructing structures.
Since Minecraft is a 3D, first-person, embodied domain and the agent's surroundings are varied and dynamic, it presents many of the same challenges as real-world robotics domains, like determining a good representation of the environment and planning over long time horizons.

\section{Competition Overview}

\label{sec:competition}

In line with its previous iteration~\citep{gussminerlneurips2019,milani2020minerl}, the MineRL 2020 Competition challenges teams to submit reproducible~\citep{houghton2020guaranteeing} training code for an agent that can solve a complex, long time-horizon task with robustness to environment domain-shift under a strict sample and computational budget. 
We describe the competition design in~\Cref{sec:comp-design}, including the mechanisms we implemented to ensure the development of robust and sample efficient learning algorithms.
The primary competition task is the MineRL \texttt{ObtainDiamond} environment, which we detail in~\Cref{sec:task}.
We summarize the rules of the competition in~\Cref{sec:rules}.
Our methodology for evaluating submitted algorithms is explained in~\Cref{sec:eval}.
To assist participants with developing their learning algorithms, we provide them with a number of important resources, including a set of reinforcement learning and imitation learning baselines, which we describe in~\Cref{sec:resources}.

\subsection{Competition Design} \label{sec:comp-design}

The MineRL Competition is designed to promote the development of robust, domain agnostic, and sample efficient algorithms for solving complex, long time-horizon tasks with sparse rewards using human priors.  
     
\paragraph{Reproducibility and Sample Efficiency.} 
 To yield sample efficient algorithms, we provide participants with the 60 million frame \texttt{MineRL-v0} human demonstration dataset~\citep{gussminerlijcai2019} of the competition task.
These samples allow the use of imitation learning techniques, which can drastically reduce the number of resources and samples required to solve complex tasks. 
Gathering expert demonstrations is practical for many RL environments, so this approach can be applied in many competitions.
 
To further ensure reproducibility and sample efficiency, we retrain participant submissions during Round 2. In addition, to directly address the problem of disproportionate and limited access to computing resources across the AI research community, we deliberately limit the hardware available for training; this further enables the democratization of AI research and the development of novel AI techniques with a low-barrier to reproduction.
    In Round 1, participants develop and train a learning algorithm on the environment using a fixed hardware and compute budget (1 NVIDIA P100 GPU for 4 days) and a fixed number of environment samples (8,000,000 frames or approximately 114 hours of game time). This compute budget was chosen as it represents an upper bound on consumer hardware; to democratize access to reinforcement and imitation learning research, constraining compute to within a reasonable consumer range is crucial. The competitors then submit their trained agent to the competition evaluator, where it is evaluated with a fixed set of holdout seeds and compared to other submissions. 
    All participants with a non-zero score progress to Round 2.
    In this round, competitors submit only their training code to the evaluator, after which it is run from scratch using the hardware, compute, and sample limits.
    This retraining procedure ensures that the submissions obey the competition limits on compute and samples.

    \begin{figure}
        \center
        \includegraphics[width=0.7\textwidth]{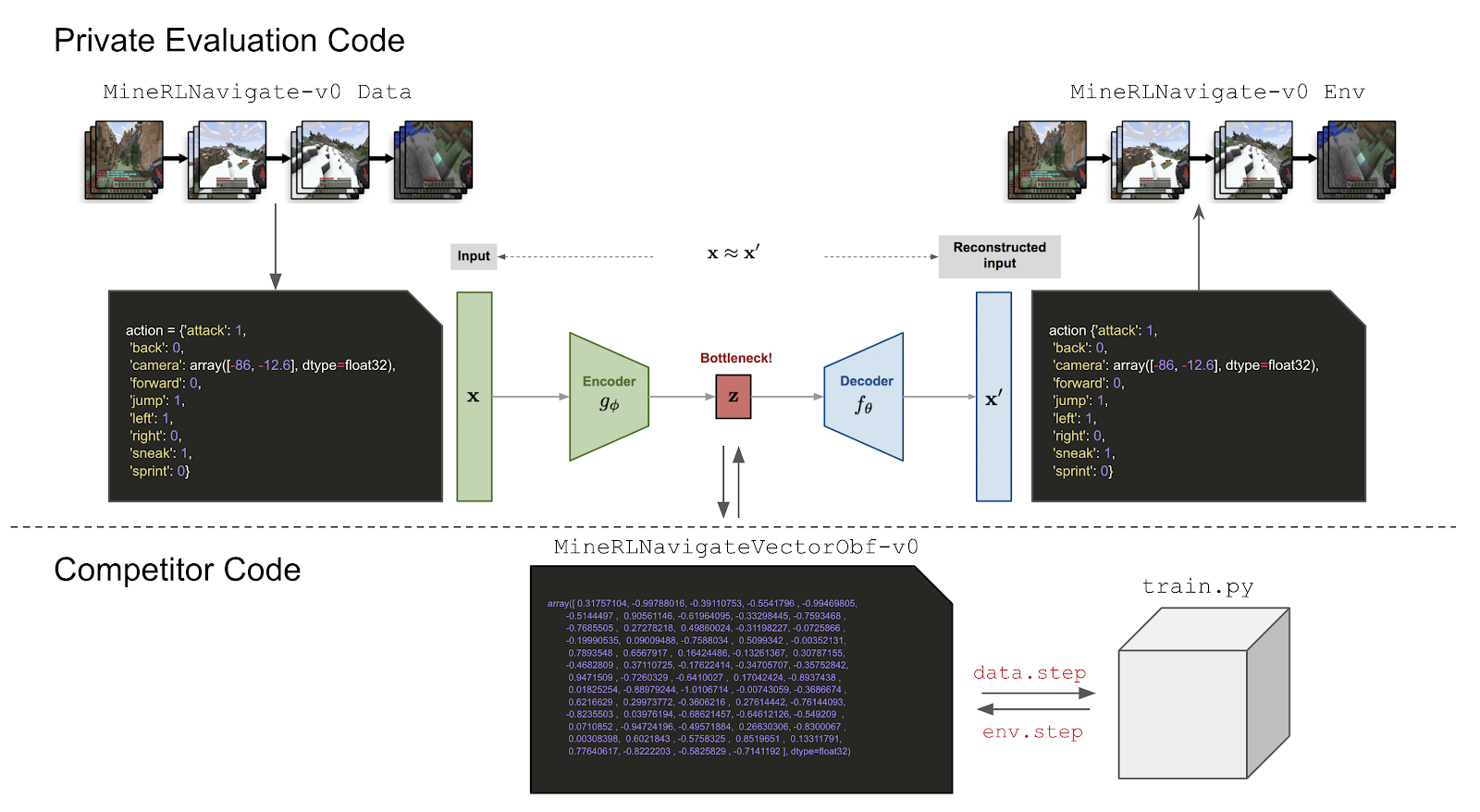}
                 \caption{\small{The action and observation space obfuscation mechanism; a randomized autoencoder trained to encode the entire mixed discrete-continuous observation and action space into a compact ball with which participants' models interact. This obfuscation prevents action shaping and feature engineering and enables domain randomization during Round 2.}}
        \label{fig:obfusc}
        \vspace{-.5cm}
    \end{figure}

\paragraph{Robustness and Domain Agnosticism.} Preventing domain specific solutions is a difficult task. 
As in the previous iteration of the MineRL Competition , during Round 2, the environment and dataset is randomized (textures are remapped, action effects are randomized, and game dynamics are changed), thus penalizing submissions which rely on domain specific strategies and feature engineering. Despite this mechanism in last year's competition, participants leveraged small, shaped subpolicies. These subpolicies were not robust to domain shifts because they depended on the semantics of the environment. 

In this year's competition, we introduce a novel obfuscation scheme that prevents domain specific hard-coding. Specifically, we learn a random, volume-preserving embedding that takes semantically-labeled actions and observations (e.g., crafting or inventory items) and obfuscates them into feature vectors. This scheme is agnostic to the environment, as algorithms trained to solve environments with this vector observation and action space can be immediately retrained against a different environment given a corresponding embedding of the new environment's action and state spaces.

Shown in \Cref{fig:obfusc}, we obtain this embedding with careful considerations of injectivity and surjectivity. 
Let $X \subset A$ be some bounded action/observation space (discrete or continuous), $P_X$ be the default sampling distribution for that space (often uniform). 
Let $Z$ be some bounded subset of $\mathbb{R}^n$ into which we wish to obfuscate $X$. 
In the MineRL Competition, $Z = [-1,1]^n$ where $n$ is the length of the obfuscated feature vector. 
Let $d_X(u,v)$ be a natural reconstruction metric for $X$ (normed difference squared for continuous $X$, and cross entropy for discrete $X$). Let $g_\theta: A \to \mathbb{R}^n$ and $f_\theta: \mathbb{R}^n \to A$ be encoder and decoder networks. We train these maps to encode the original observation/action space $X$ respecting the bounds of the space by minimizing reconstruction loss while maintaining that $g_\theta(X) \subset Z$ and $f_\theta(Z) \subset X$; that is, we define our reconstruction loss as 
\begin{equation*}
    \begin{aligned}
    L_X(\theta) = \mathbb{E}_{y \sim P_{X}}[d_X(f_\theta(g_\theta(y)), y) &+ \texttt{Hinge}(g_\theta(y), Z)] 
    \\&+ \mathbb{E}_{z \sim \mathsf{Unif}(Z)}[\texttt{Hinge}(f_\theta(z), X)]  
    \end{aligned}
\end{equation*}
where $\texttt{Hinge}(u, V)$ is a hinge loss which is zero when $u$ is in $V$ and piecewise linear, increasing otherwise. For example, for $Z$, a simple box space, $\texttt{Hinge}(z, [-1,1]^n) = \sum_{i=1}^n \texttt{ReLu}(|z_i| -1)$. This loss accomplishes two goals: when the competitors use $Z$ as an action space, they are approximately guaranteed to have a valid action $f_\theta(z)$ in $X$ when sampling within the bounds of $Z$. Furthermore, the reconstruction loss ensures that all possible actions in $X$ can be taken by finding a point in $Z$. Likewise, when using $Z$ as an observation space, all observations in the unobfuscated action space $X$ are approximately guaranteed to be contained inside of $Z$.
Therefore, competitors can appropriately normalize their observations in $Z$. It is important that $Z$ is of the same or higher dimensionality than $X$ so there is certain to be a minimizer $\theta^*$. In the MineRL Competition, both action and observation space embeddings were trained with $\text{dim}(Z) = 64$ to an error of at most $\texttt{1e-12}$; we chose this space as it was of high enough dimension to embed the observation and action space but of low enough dimension that reinforcement learning algorithms converge within the compute budget.

\subsection{Task}
\label{sec:task}
 The primary task of the competition is \texttt{ObtainDiamond}. 
Agents begin at a random position on a randomly-generated Minecraft map with no items in their inventory. 
Completing the task consists of controlling an embodied agent to obtain a single diamond, which can only be accomplished by navigating the complex item hierarchy of Minecraft. 
The learning algorithm has direct access to a $64$x$64$ pixel point-of-view observation from the perspective of the embodied agent, as well as a set of discrete observations of the agent's inventory for every item required for obtaining a diamond.  The action space is the Cartesian product of continuous view adjustment (turning and pitching), binary movement commands (left/right, forward/backward), and discrete actions for placing blocks, crafting items, smelting items, and mining/hitting enemies.  
An agent receives reward once per episode for reaching a set of milestones of increasing difficulty that form a set of prerequisites for the full task.
~\Cref{table:rew} depicts the full reward structure.

Progress towards solving the \texttt{ObtainDiamond} environment under strict sample complexity constraints lends itself to the development of sample-efficient--and therefore more computationally accessible--sequential decision-making algorithms. 
In particular, because we maintain multiple versions of the dataset and environment for development, validation, and evaluation, it is difficult to engineer domain-specific solutions to the competition challenge. 
The best performing techniques must explicitly implement strategies that efficiently leverage human priors across general domains.

\subsection{Rules}\label{sec:rules}
Due to the unique competition paradigm, we provide a strict set of rules to ensure high-quality submissions. 
We prohibit teams from manually engineering the reward function, action space, and observation space. 
For example, we permit curiosity rewards but not bonus rewards for encountering specific objects; we allow a learned hierarchical controller but not one that switches between policies based on manually-specified conditions; we allow agents to act every even-numbered timestep based on the previous two observations but prohibit the application of manually specified edge detectors to the observation. 
Furthermore, we require that competitors' code make no semantic reference to the environment.
To encourage reproducible submissions, we require entries to the competition to be open: teams must reveal most details of their method, including the source code.
Moreover, the competition features two sets of tracks, each of which have distinct rules.
The \rlil Track features methods that leverage both samples from the environment and human demonstrations, while algorithms in the \ilonly Track must only use imitation learning with no access to the environment --- except for during evaluation.

\subsection{Evaluation}
\label{sec:eval}
\paragraph{Submission Platform.}
The submissions for the competition were evaluated using AIcrowd, which has been used in numerous RL benchmarks~\citep{kidzinski20f8learning,juliani2019obstacle, mohanty2020flatland}, and allows for the much needed flexibility when designing complex benchmarks. 
Participating teams independently develop their solutions using Git repositories provided by AIcrowd. 
The repositories include simple configurations specifying the expected software runtime. 
They also have a prescribed structure to enable clear specifications of code entry points for different phases of evaluation.
Participants submit to the benchmark by releasing Git tags, which trigger the evaluation workflow. 
The workflow then builds a Docker image with the submitted code repository, which is automatically orchestrated on a scalable Kubernetes cluster. 
During evaluation, the evaluators provide real-time feedback on the progress of the evaluation and submission-specific metrics to the participants. If a submission fails, participants can debug their submission using the readily-available logs. To avoid data leak, care is taken to ensure that logs are not made available for sensitive phases of the evaluation. On successful completion of the evaluation workflow, the evaluators update the scores, and any generated assets on the competition leaderboard. 

\begin{wraptable}{r}{0.4 \textwidth}
    \centering
    \vspace{-12 pt}
    \scalebox{.6}{
    \begin{tabular}{ll|ll} 
        {Milestone} & {Reward} & {Milestone} & {Reward} \\
        \midrule
        log & 1                & furnace & 32  \\
        planks & 2             & stone\_pickaxe & 32  \\
        stick & 4              & iron\_ore & 64  \\
        crafting\_table & 4    & iron\_ingot & 128  \\
        wooden\_pickaxe & 8    & iron\_pickaxe &  256 \\
        stone & 16             & diamond & 1024 
    \end{tabular}
    }
    \vspace{-0.3cm}
    \caption{
        \small Rewards for sub-goals and main goal (diamond) for \texttt{Obtain Diamond}. }

    \label{table:rew}
\end{wraptable} 

\paragraph{Metrics.}
\label{sec:metrics}

When models are submitted in Round 1 and after they are trained by organizers in Round 2, participants are evaluated on the average score of their model over 200 episodes.
Scores are computed as the sum of the milestone rewards (shown in~\Cref{table:rew}) achieved by the agent in a given episode. 
 Ties are broken by the number of episodes required to achieve the last milestone.

\subsection{Resources}
\label{sec:resources}
 In addition to providing the \texttt{MineRL-v0}  dataset~\citep{gussminerlijcai2019}, we give participants an open-source Github repository with starting code, including an OpenAI Gym template interface, a data-loader, a Docker container, and the code for the solutions created by last year's top participants.
 We also provide participants with a set of four state-of-the-art baselines that they could readily submit.
Implemented by the organizers from Preferred Networks, these baselines consist of Soft-Q Imitation Learning, (SQIL) ~\citep{DBLP:conf/iclr/ReddyDL20}, Deep-Q From Demonstrations (DQfD) ~\citep{hester2018deep}, Rainbow Deep-Q Networks (Rainbow) ~\citep{hessel2018rainbow}, and Prioritized Dueling Double Deep Q-Networks (PDDDQN) ~\citep{schaul2015prioritized,van2016deep,wang2016dueling}. 
The baselines all used K-means clustering~\citep{macqueen1967some,lloyd1982} to discretize the action space.

\section{Solutions}

\begin{figure}
  \vspace{-.3cm}
\centering
\includegraphics[width=0.7\textwidth]{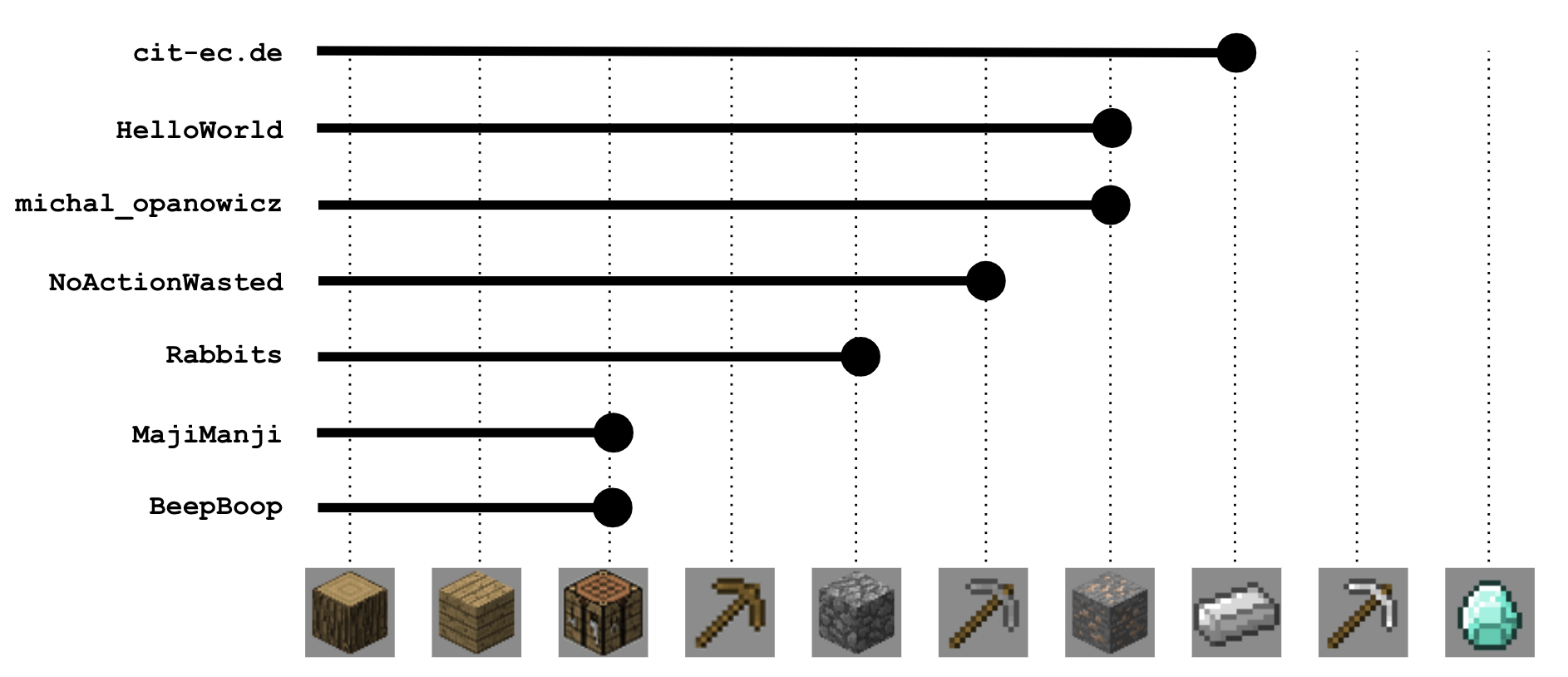}
\vspace{-.5cm}
\caption{\small{Maximum item score for each team over the evaluation episodes in Round 2.}}
\vspace{-0.2cm}
\label{fig:progress}
\end{figure}

\begin{table}
  \small
  \centering
    \begin{tabular}{|c|r||c|r||c|r|}
       \hline
       \multicolumn{2}{|c||}{\textbf{Baselines}} &
       \multicolumn{2}{c||}{\textbf{Round 1}} &
       \multicolumn{2}{c|}{\textbf{Round 2}}\\
       \hline 
       \textbf{Name} & \textbf{Score} & \textbf{Team Name} & \textbf{Score} & \textbf{Team Name} & \textbf{Score} \\
       \hline 
       \team{SQIL} & 2.94 & \team{HelloWorld} & 19.84 & \team{cit-ec.de} & 72.51\\
       \team{DQFD} & 2.39 & \team{NoActionWasted} & 16.48 & \team{HelloWorld} & 39.55\\
       \team{Rainbow} & 0.42 & \team{michal\_opanowicz} & 9.29 & \team{michal\_opanowicz}  & 13.29\\
       \team{PDDDQN} & 0.11 & \team{CU-SF} & 6.47 & \team{NoActionWasted} & 12.79\\
       &  & \team{cit-ec.de} & 6.40 & \team{Rabbits} & 5.16\\
       &  & \team{NuclearWeapon} & 4.34 & \team{MajiManji} & 2.49\\ 
       &  & \team{murarinCraft} & 3.61 & \team{BeepBoop} & 1.97\\
       &  & \team{RL4LYFE} & 3.39 &  & \\
       &  & \team{porcupines} & 3.35 &  & \\ 
       \hline 
            \end{tabular}
    \caption{\small{Scores of the baselines (left) and the best-performing submissions from Round 1 (middle) and Round 2 (right).}}
    \vspace{-0.7cm}
    \label{tab:scores2020}
\end{table}
\label{sec:solutions}
\label{sec:solutions}
\label{sec:solutions}
We provide an overview of the submissions made by the participants of our competition.
In~\Cref{sec:performance-overview}, we describe the performance of the submissions  and how they compare to the performance of submissions to the 2019 competition. 
The remaining sections summarize the techniques used by the competitors in their submissions.

\subsection{Submission Performance Overview}
\label{sec:performance-overview}
 The conditions surrounding the competition and the changes to the competition itself proved to make the competition more challenging for the competitors compared to last year.
Although this year's competition enjoyed participation from more teams than last year's ($95$ vs. $47$), there were fewer submissions overall ($513$ vs. $662$).
 We believe that this decrease in submissions is in part due to the global pandemic.
Teams still performed well: in Round 1, $36$ teams achieved a non-zero score, and $17$ of these teams outperformed the best-performing baseline, SQIL. 
In Round 2, seven teams achieved a non-zero score and some teams performed even better than they did in Round 1.

~\Cref{tab:scores2020} shows the scores of the best-performing submissions from both rounds of the 2020 competition.
The average scores of the top nine competitors in the 2020 competition were $8.13$ for Round 1 and $16.42$ for Round 2.
 In contrast, the average scores of the top nine competitors in the 2019 competition were $31.77$ for Round 1 and $25.84$ for Round 2.
Surprisingly, the standard deviation of the top nine scores from Round 1 of 2020 was smaller than the standard deviation of the top nine scores from Round 1 of 2019 ($5.72$ and $10.22$, respectively). 
This finding may be due to the overlap of techniques used by the competitors: at least four of the top nine competitors in the 2020 competition leveraged a similar K-means clustering of the action space.
~\Cref{fig:progress} depicts the maximum item score for each team over the evaluation episodes in Round 2.
Although no team obtained a diamond, many of the top teams progressed quite far along the item hierarchy.

\vspace{-10pt}
\subsection{Team 1: \team{cit-ec.de}}

\begin{figure}[h!]
  \centering
  \includegraphics[width=12.1cm]{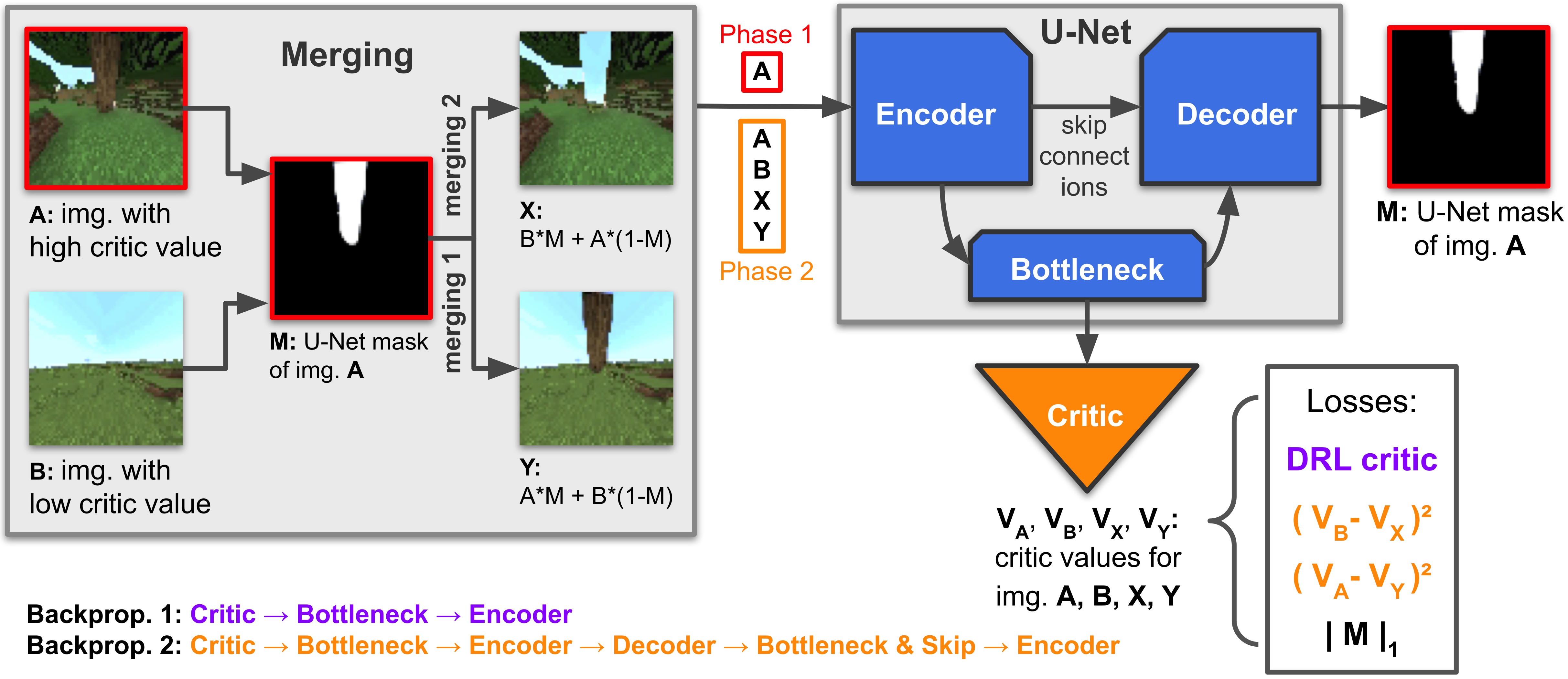} 
  \vspace{2mm}
  \includegraphics[width=12.1cm]{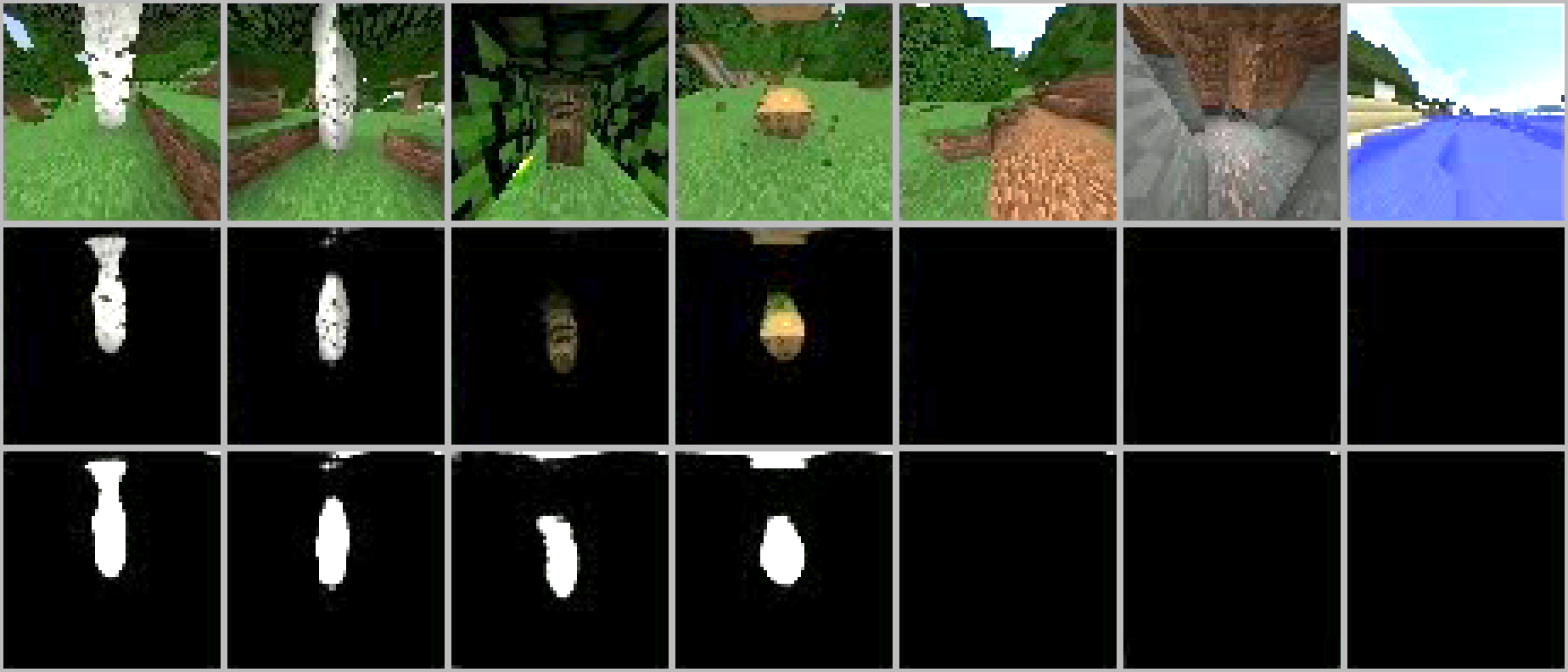}
     \caption{\small{
  \team{cit-ec.de}'s method.
  \textbf{Top:} First phase (highlighted in red): Image \textbf{A} (high critic value) passes through the U-Net, forming a mask \textbf{M}. Second phase: the mask \textbf{M} is used to merge image \textbf{A} (high critic value) with image \textbf{B} (low critic value) resulting in image \textbf{X} (masked parts of \textbf{A} replaced with \textbf{B}) and image \textbf{Y} (masked parts of \textbf{A} injected in \textbf{B}). Images \textbf{A}, \textbf{B}, \textbf{X}, and \textbf{Y} are then passed through the encoder and critic. The losses penalize differences in critic values for image pairs \textbf{A} : \textbf{Y}, and \textbf{B} : \textbf{X}. Mask-size loss prevents a trivial solution when \textbf{M} takes the full image.
\textbf{Bottom:} Segmentation results: The U-Net model learns to segment tree trunks without any label information but only from reward signals. It generalizes well between different positive and negative reward scenarios. The first row shows the input images, the second row shows the masked segments of the input images, and the third row shows the U-Net generated masks.}}
  \label{fig-unet}
  \label{fig-results}
  \end{figure}

  Overall, team \team{cit-ec.de} placed fifth in Round 1 (score of $6.40$) and first in Round 2 (score of $72.510$).
They also placed first in the \rlil track. 
In Round 1, they competed in the \ilonly track, but they switched to the \rlil Track in Round 2.
Inspired by the recent success of cognitive science research~\citep{melnik2018world,konig2018embodied} and its applications in artificial intelligence systems~\citep{melnik2019combining,konen2019biologically,bach2020learn,melnik2019modularization,harter2020solving,schilling2018approach}, their approach aims to learn to detect object-centric representations from pixels~\citep{simonyan2013deep} using rewarding signals of interaction with the environment.

Depicted in~\Cref{fig-unet} (top), they train a U-Net model to generate masks over reward-related objects in images. 
This approach enables the training of the U-Net model without explicit label information. 
Instead, they perform this training in a contrastive fashion with image pairs using an adversarial scheme employing the critic score gradient with respect to the mask operation. 
The pair consists of two images, where the first has a high and the second a low critic value. 
Training with such pairs enables the U-Net to produce masks that decrease the critic value in the first image and increase the critic value in the second image when transferring pixels in the masked segment from the first to the second image. 
The critic learns to estimate the expected-reward value of an image observation using experience replay buffer collected from human player demonstrations. 
As shown in~\Cref{fig-results} (bottom), this approach of training the U-Net model showed encouraging results of segmentation of rewarding objects in the competition~\citep{melnik2021critic}.

\vspace{-6pt}
\subsection{Team 2: \team{HelloWorld}}
\begin{wrapfigure}{l}{0.36\textwidth}
\centering
\vspace{-15pt}
\includegraphics[scale=0.25]{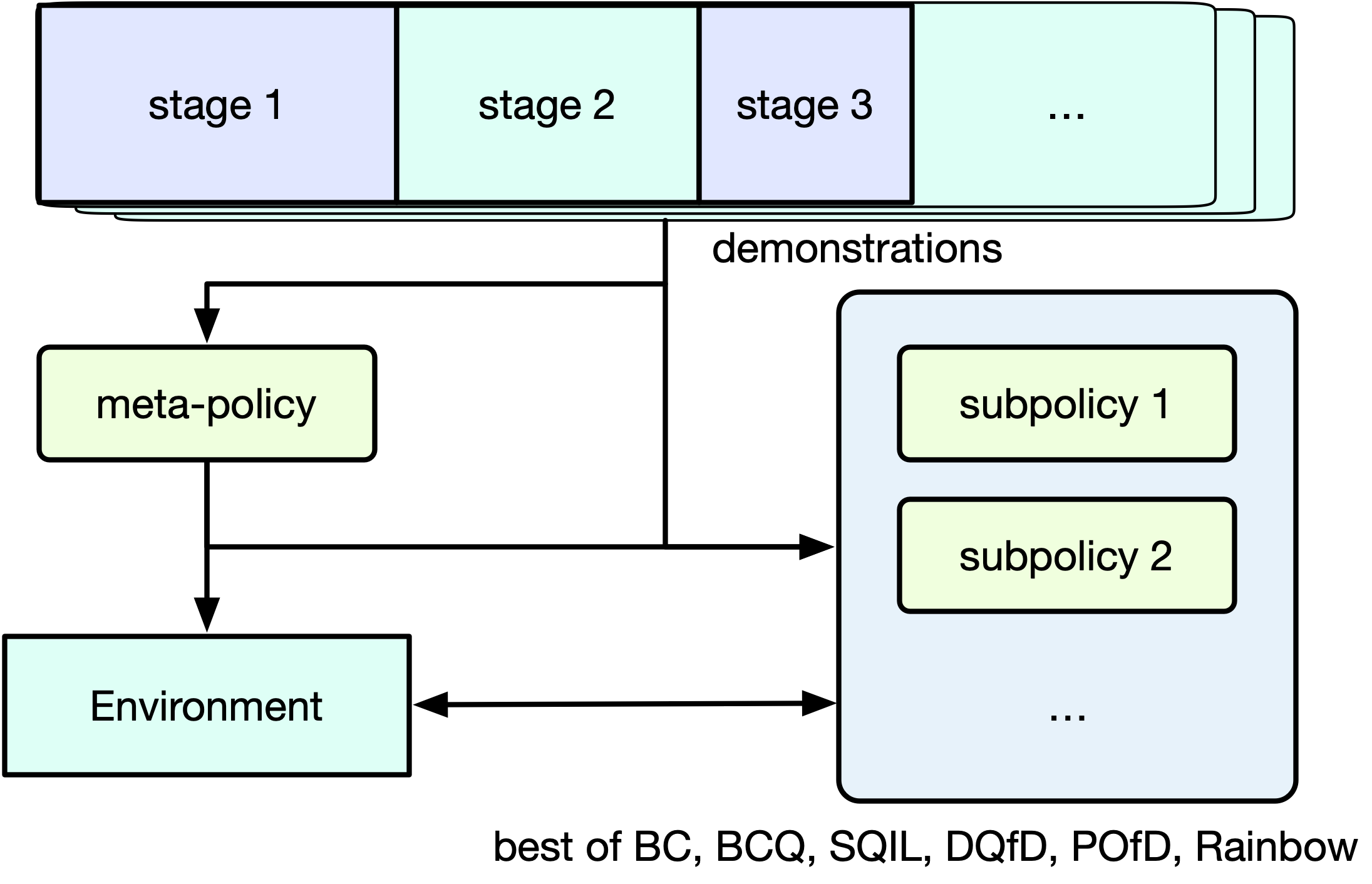}
\vspace{-10pt}
\caption{\team{HelloWorld}'s method.}
\vspace{-1cm}
\label{fig:helloworld_framework}
 \end{wrapfigure}
Team \team{HelloWorld} placed first in Round 1 (score of 19.840) and second in Round 2 (score of $39.55$).
They competed in the \rlil track.
Shown in Figure~\ref{fig:helloworld_framework}, their approach splits each episode into several stages based on accumulated rewards.
They train one policy for each stage of the episode to enable each subpolicy to capture different information towards different goals.
They train a meta-policy to select a subpolicy for execution at each timestep. 
For each subpolicy, they select the best algorithm from all of the tested ones, including behavioral cloning, BCQ~\citep{fujimoto2019off}, SQIL, DQfD, POfD~\citep{kang2018policy}, and Rainbow.
They find that SQIL incorporated with self-imitation avoids performance drop during training.
Finally, their solution consists of other key components to make their approach more robust and generalizable, including pretraining agents on simpler environments, leveraging auxiliary tasks (e.g., predicting future state), and applying state augmentation~\citep{yarats2021image}.
\subsection{Team 3: \team{michal\_opanowicz}}
\begin{wrapfigure}{r}{0.5\textwidth}
  \centering
  \includegraphics[width=6.5cm]{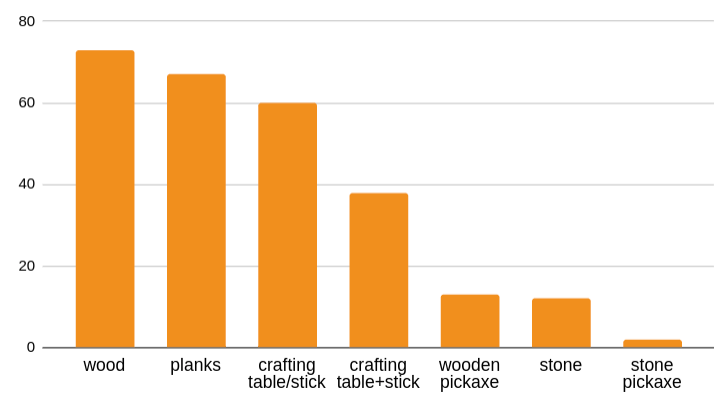}
  \caption{The percentage of runs in which \team{michal\_opanowicz}'s agent collected a given item.}
  \label{fig:michal_perf}
  \vspace{-5pt}
  \end{wrapfigure}
Team \team{michal\_opanowicz} achieved third place in Round 1 (score of 9.290) and Round 2 (score of 13.290).
However, this team achieved first place overall in the \ilonly track.
They use imitation learning, in which the problem is framed as a classification task where the agent predicts the human player's action at each environment step. 
To produce discrete labels for this task, they quantize the actions using K-means with $120$ clusters. 
During evaluation, they randomly sample the actions from the cluster means with the network's prediction used as a probability distribution.

For visual processing, they use the ResNet~\citep{DBLP:journals/corr/HeZRS15} architecture with FixUp initialization~\citep{DBLP:journals/corr/abs-1901-09321}, that was proposed for this task in a MineRL 2019 submission~\citep{amiranashvili2020scaling}. 
They process non-visual observations with a fully connected layer with ReLU activation and concatenate it with the ResNet outputs. 
They are then processed by a LSTM~\citep{10.1162/neco.1997.9.8.1735} and two fully connected layers to produce the final prediction.
Notably, they add the LSTM inputs to its outputs to form a residual connection similar to ResNets. 
They train the network on $100$-step sequences of observation-action pairs, with the final LSTM state in a sequence used as the initial state for the next sequence.
~\Cref{fig:michal_perf} shows an analysis of their algorithm's performance.
In most runs, their agent obtains either a crafting table or a stick.

\subsection{Team 4: \team{NoActionWasted}}
\begin{wrapfigure}{r}{0.5\textwidth}
\centering
\vspace{-1cm}
\includegraphics[width=0.5\textwidth]{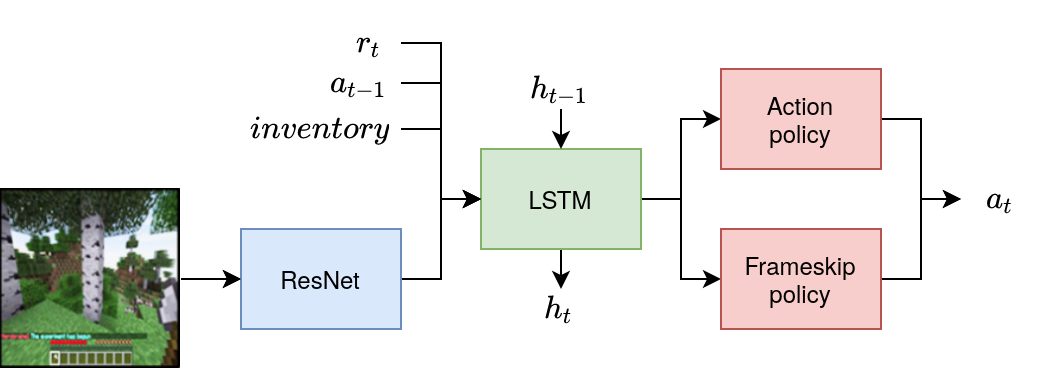}
\caption{\small{\team{NoActionWasted}'s architecture.}}
\label{fig:networkarchitecture}
 \end{wrapfigure}
Team \team{NoActionWasted} competed in the \ilonly track. 
Overall, they achieved second place in Round 1 and fourth in Round 2 (scores of 16.48 and 12.79, respectively). 
Depicted in Figure~\ref{fig:networkarchitecture}, their system consists of a ResNet-LSTM network~\citep{espeholt2018impala} trained to predict human actions. 
They found that directly training agents on the obfuscated actions was not successful, so they discretized the action space into $150$ clusters with K-means.  This step was crucial to obtain good initial behavior. 
When using a lower amount of clusters, some rare but important actions are not represented. 
Instead of fixing the frame-skip parameter, they train the network to predict it as an action parameter , which provides significant benefit for agents in the Minecraft domain, as many tasks require repeating the same action multiple times, and it reduces the perceived episode length.

They filter the dataset to only include successful games, which consistently improved the performance.
They fine-tune the system with IMPALA~\citep{espeholt2018impala}.
Following their submission last year~\citep{scheller2020sample}, they pair this algorithm with extensive experience replay~\citep{lin1992self}, clipping advantages to promote exploitation of good behavior, and CLEAR~\citep{rolnick2018experience} to combat catastrophic forgetting. 
Crucially, they find that large batch sizes were crucial for stability during the RL fine tuning.
  
\subsection{Team 5: \team{Rabbits}}
Team Rabbits achieved fifth place in Round 2 (score of $5.16$).
Although they competed in both tracks, their highest-performing submission was from their \rlil track submission.  For their approach in both tracks, they split the overall task into $10$ subtasks based on reward, transforming it into a hierarchical learning problem. 
They use a task-classification network to determine which of the $10$ reward stages the agent is currently in. The task-classification network only takes the state vector as input. 
To learn the subtasks, they use $10$ separate Q networks.
For the \ilonly track submission(s), they employ Regularized Behavior Cloning (RBC)~\citep{piot2014boosted} on the demonstration data. 
They train each Q network separately, stopping when the TD error no longer decreases.
For their \rlil submission(s), they apply RBC as the starting point for RL. 
The RL algorithm they use is a variant of SQIL: they set the reward in expert replay buffer to be half of the reward of the corresponding task, except for the steps that obtain the sparse reward.

\subsection{Team 6: \team{MajiManji}}
\begin{wrapfigure}{r}{0.4\textwidth}
  \vspace{-1cm}
  \centering
  \includegraphics[width=0.4\textwidth]{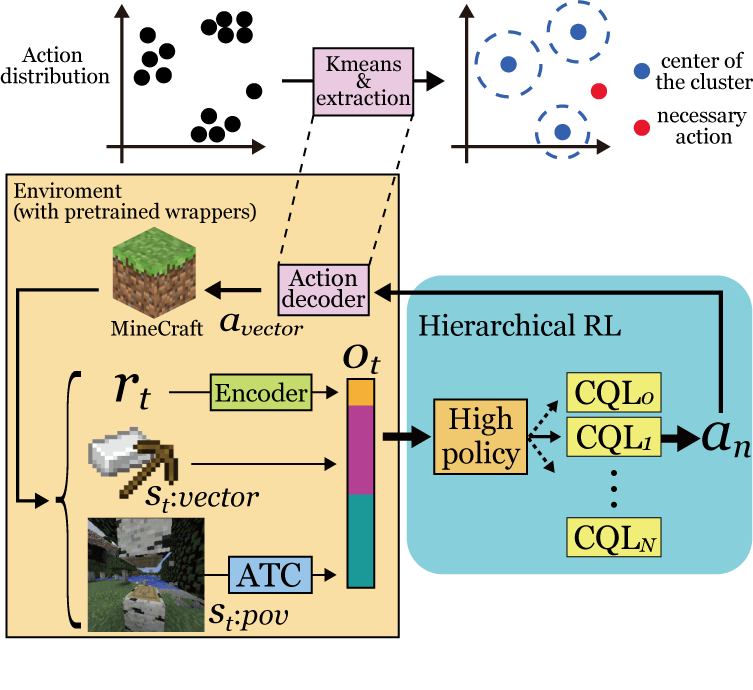}
  \caption{Overview of \team{MajiManji}'s architecture.}
  \vspace{-0.7cm}
  \label{fig:overview}
  \end{wrapfigure}

Team \team{MajiManji} achieved sixth place in Round 2 (score: 5.16). 
They participated in the \rlil track.
Their approach uses hierarchical offline reinforcement learning.
Outlined in~\Cref{fig:overview}, they decompose the high-level task of mining a diamond into a set of subtasks, and train a separate CQL agent~\citep{kumar2020conservative} for each subtask.
To determine these subtasks, they use a label encoder that maps cumulative reward to a label.
The high-level policy leverages the subtask label to select the low-level policies.
 In the environments ending with ``VectorObf'', which have both POV images and vectors as states, they use ATC~\citep{stooke2020decoupling} to extract features only from the POV images.

They apply a K-means algorithm to the action space for each subtask. 
In addition, they construct a ``necessary action space''.
An action is added to this action space when the total number of times it is performed is few, but it is performed in most episodes.  
 Since the low-level policies may not learn to do these necessary actions, the high-level policy chooses a necessary action randomly with low probability $ \epsilon $.
In the future, they plan to find how to weight important demonstrations with no reward and few samples.

\subsection{Team 7: \team{BeepBoop}}
Team \team{BeepBoop} achieved twelfth place in Round 1 (score: 3.110) and seventh in Round 2 (score: 1.970).
In Round 1, they participated in the \rlil track, where they simply used the SQIL baseline.
For Round 2, they participated in both tracks.
Their \rlil track submission was designed with the goal of improving the DQfD baseline.
First, they improved data preparation: since their agent never reached the later stages of the challenge, they only use the data from the earlier stages of the \texttt{ObtainDiamond} and \texttt{ObtainIronPickaxe} environments, as measured by cumulative rewards. 
They also included the full TreeChop dataset.
Second, they focused on improving the action space representation.
They use K-means clustering with a large number of clusters to cluster the action space to ensure that the agent could access all necessary actions.
For their \ilonly track submission, they used behavioral cloning with the ResNet-50 network architecture~\citep{DBLP:journals/corr/HeZRS15}.
For the labels, they used the centroids of K-means clustered obfuscated actions. 
They apply the same data preparation and action space modification techniques as in the RL+IL track submission.

\section{Discussion}
\label{sec:discussion}

  Promoting the development of sample-efficient, domain agnostic, and reproducible RL algorithms is crucial in translating the research advances of RL into complex, real-world settings. In this section, we discuss the extent to which the design principles behind the MineRL competition have accomplished this goal.

\paragraph{Successes.} As the results of~\Cref{sec:solutions} illustrate, the submissions to the competition are general and successful in the face of the competition constraints. Competitors submitted sample-efficient algorithms that made substantial progress towards the difficult \texttt{ObtainDiamond} task with a mere 8,000,000 frames from the environment. 
Further, despite the domain randomization in Round 2, many of the top algorithms from Round 1 also achieved impressive scores in Round 2. 
The submissions that performed well in Round 2 were those which were robust to domain-shift, and because the organizers retrained submissions from scratch in Round 2, those successful submissions were also certifiably reproducible and resource-efficient. 
Unlike the previous iteration of the MineRL Competition, the addition of the action/observation obfuscation technique completely prevented teams from action shaping and feature engineering, leading to the emergence of a very common action-simplification technique: K-means clustering of the expert action space.

\paragraph{Limitations.} 
  In its current form, the competition is limited to the Minecraft domain.
   Although domain randomization and action space obfuscation prevent the introduction of some inductive biases into the participants' algorithms, competitors still expect to be faced with the broader game mechanics of Minecraft. 
      This domain knowledge ultimately manifests itself in the particular algorithmic structure of participants' submissions (e.g., in the choice of hierarchical methods).
      A natural next step to promote more domain agnostic methods is to expand the scope of the competition to include a training step against a different domain using the autoencoder-based technique proposed in~\Cref{sec:comp-design}.
      This step would ensure submissions developed for the \texttt{ObtainDiamond} task could succeed in an auxiliary domain.
   An additional limitation of the competition design is that the obfuscation of the observation space does not include the pixel observation subspace. Although this choice decreases the scope of research outcomes in the competition, it directly promotes research towards algorithms that learn from pixels, a current and crucial challenge in deep RL.

\section{Related Work}
\label{sec:related}
 Previous competitions~\citep{juliani2019obstacle} focused on a variety of aspects of RL, such as the multi-agent setting~\citep{gao2019skynet,perez2019multi,mohanty2020flatland} or practical applications~\citep{marot2020l2rpn}. 
 In most of these competitions, the focus is not on generalization.
Instead, the goal is to develop a trained agent that performs well on a given domain.  Consequently, winning submissions often relied on hand-engineered features and stemmed from the use of large amounts of computational resources to optimize the submission for the specific task.
Competitions that have promoted the development of general RL algorithms do not perform action or observation obfuscation and either focus on generalization across a variety of tasks with shared objects, textures, and actions~\citep{nichol2018gotta} or utilize the approach of having hold-out test tasks~\citep{cartoni2020real}.

    To our knowledge, no previous competition has explicitly encouraged the use of imitation learning alongside RL.
Previous imitation learning competitions~\citep{diodato2019winning} concentrate on the prediction setting (e.g., predicting a vehicle's speed given sensor inputs), which is reflected in their evaluation metrics.
However, our evaluation metrics are more reflective of the sequential decision-making setting in which our competition takes place.
  Although some top solutions to previous competitions leveraged imitation learning~\citep{meisheri2019accelerating}, the use of imitation learning alongside reinforcement learning was not explicitly promoted.
In contrast, we encourage the use of imitation learning by providing participants with a large dataset of human demonstrations and introducing a second track to the competition where competitors must only use the dataset to train their algorithms.

\section{Conclusion}\label{sec:conclusion}
We ran the 2020 MineRL Competition on Sample Efficient Reinforcement Learning Using Human Priors in order to promote the development of general, sample efficient reinforcement learning and imitation learning algorithms.
We described the competition, highlighting changes to the rules and structure from the 2019 version of the competition.
We summarized the performance of the submissions and contrasted this performance with the performance from last year.
We described the approaches of the top seven teams from Round 2.
We concluded by discussing the benefits and limitations of our approach.

\acks{
   
   We thank AIcrowd for hosting the competition evaluator and providing tireless hours of support in ensuring that competitors could submit their solutions. We especially thank Shivam Khandelwal for his help in developing the competition starter-kit and providing constant assistance to the organizers and the participants during the competition. 
   We would like to thank Ansii Kanervisto for his continual and detailed responses in the competition Discord. In addition, we would like to acknowledge our sponsor Preferred Networks for providing a rich set of baselines in their new framework PFRL.     We thank Microsoft Research and the Artificial Intelligence Journal for their generous sponsorship of competition compute (for retraining and evaluation), of compute grants enabling the participation of underrepresented groups, and of NeurIPS registrations for competitors. Finally, we would like to acknowledge Microsoft Research and NVIDIA for providing prizes for the competition.
   
}

\bibliography{minerl_deduped.bib}

\begin{thebibliography}{66}
\providecommand{\natexlab}[1]{#1}
\providecommand{\url}[1]{\texttt{#1}}
\expandafter\ifx\csname urlstyle\endcsname\relax
  \providecommand{\doi}[1]{doi: #1}\else
  \providecommand{\doi}{doi: \begingroup \urlstyle{rm}\Url}\fi

\bibitem[Amiranashvili et~al.(2020)Amiranashvili, Dorka, Burgard, Koltun, and
  Brox]{amiranashvili2020scaling}
Artemij Amiranashvili, Nicolai Dorka, Wolfram Burgard, Vladlen Koltun, and
  Thomas Brox.
\newblock Scaling imitation learning in minecraft.
\newblock \emph{arXiv preprint arXiv:2007.02701}, 2020.

\bibitem[Amodei and Hernandez(2018)]{amodei_hednandez_2018}
Dario Amodei and Danny Hernandez.
\newblock https://blog.openai.com/ai-and-compute/, May 2018.
\newblock URL \url{https://blog.openai.com/ai-and-compute/}.

\bibitem[Bach et~al.(2020)Bach, Melnik, Schilling, Korthals, and
  Ritter]{bach2020learn}
Nicolas Bach, Andrew Melnik, Malte Schilling, Timo Korthals, and Helge Ritter.
\newblock Learn to move through a combination of policy gradient algorithms:
  Ddpg, d4pg, and td3.
\newblock In \emph{International Conference on Machine Learning, Optimization,
  and Data Science}, pages 631--644. Springer, 2020.

\bibitem[Bellemare et~al.(2013)Bellemare, Naddaf, Veness, and
  Bowling]{bellemare2013arcade}
Marc~G Bellemare, Yavar Naddaf, Joel Veness, and Michael Bowling.
\newblock The arcade learning environment: An evaluation platform for general
  agents.
\newblock \emph{Journal of Artificial Intelligence Research}, 47:\penalty0
  253--279, 2013.

\bibitem[Berner et~al.(2019)Berner, Brockman, Chan, Cheung, D{{e}}biak,
  Dennison, Farhi, Fischer, Hashme, Hesse, et~al.]{berner2019dota}
Christopher Berner, Greg Brockman, Brooke Chan, Vicki Cheung, Przemyslaw
  D{{e}}biak, Christy Dennison, David Farhi, Quirin Fischer, Shariq Hashme,
  Chris Hesse, et~al.
\newblock Dota 2 with large scale deep reinforcement learning.
\newblock \emph{arXiv preprint arXiv:1912.06680}, 2019.

\bibitem[Buckman et~al.(2018)Buckman, Hafner, Tucker, Brevdo, and
  Lee]{buckman2018sample}
Jacob Buckman, Danijar Hafner, George Tucker, Eugene Brevdo, and Honglak Lee.
\newblock Sample-efficient reinforcement learning with stochastic ensemble
  value expansion.
\newblock \emph{arXiv preprint arXiv:1807.01675}, 2018.

\bibitem[Cartoni et~al.(2020)Cartoni, Mannella, Santucci, Triesch, Rueckert,
  and Baldassarre]{cartoni2020real}
Emilio Cartoni, Francesco Mannella, Vieri~Giuliano Santucci, Jochen Triesch,
  Elmar Rueckert, and Gianluca Baldassarre.
\newblock Real-2019: Robot open-ended autonomous learning competition.
\newblock In \emph{NeurIPS 2019 Competition and Demonstration Track}, pages
  142--152. PMLR, 2020.

\bibitem[Cobbe et~al.(2019)Cobbe, Klimov, Hesse, Kim, and
  Schulman]{cobbe2019quantifying}
Karl Cobbe, Oleg Klimov, Chris Hesse, Taehoon Kim, and John Schulman.
\newblock Quantifying generalization in reinforcement learning.
\newblock In \emph{International Conference on Machine Learning}, pages
  1282--1289, 2019.

\bibitem[Diodato et~al.(2019)Diodato, Li, Goyal, and Drori]{diodato2019winning}
Michael Diodato, Yu~Li, Manik Goyal, and Iddo Drori.
\newblock Winning the {ICCV} 2019 learning to drive challenge.
\newblock \emph{arXiv preprint arXiv:1910.10318}, 2019.

\bibitem[Dubey et~al.(2018)Dubey, Agrawal, Pathak, Griffiths, and
  Efros]{dubey2018investigating}
Rachit Dubey, Pulkit Agrawal, Deepak Pathak, Thomas~L Griffiths, and Alexei~A
  Efros.
\newblock Investigating human priors for playing video games.
\newblock \emph{arXiv preprint arXiv:1802.10217}, 2018.

\bibitem[Espeholt et~al.(2018)Espeholt, Soyer, Munos, Simonyan, Mnih, Ward,
  Doron, Firoiu, Harley, Dunning, et~al.]{espeholt2018impala}
Lasse Espeholt, Hubert Soyer, Remi Munos, Karen Simonyan, Vlad Mnih, Tom Ward,
  Yotam Doron, Vlad Firoiu, Tim Harley, Iain Dunning, et~al.
\newblock Impala: Scalable distributed deep-rl with importance weighted
  actor-learner architectures.
\newblock In \emph{International Conference on Machine Learning}, pages
  1407--1416, 2018.

\bibitem[Franke et~al.(2020)Franke, K{\"o}hler, Biedenkapp, and
  Hutter]{franke2020sample}
J{\"o}rg~KH Franke, Gregor K{\"o}hler, Andr{\'e} Biedenkapp, and Frank Hutter.
\newblock Sample-efficient automated deep reinforcement learning.
\newblock \emph{arXiv preprint arXiv:2009.01555}, 2020.

\bibitem[Fujimoto et~al.(2019)Fujimoto, Meger, and Precup]{fujimoto2019off}
Scott Fujimoto, David Meger, and Doina Precup.
\newblock Off-policy deep reinforcement learning without exploration.
\newblock In \emph{International Conference on Machine Learning}, pages
  2052--2062. PMLR, 2019.

\bibitem[Gao et~al.(2019)Gao, Hernandez-Leal, Kartal, and
  Taylor]{gao2019skynet}
Chao Gao, Pablo Hernandez-Leal, Bilal Kartal, and Matthew~E Taylor.
\newblock Skynet: A top deep rl agent in the inaugural pommerman team
  competition.
\newblock \emph{arXiv preprint arXiv:1905.01360}, 2019.

\bibitem[Guss et~al.(2019)Guss, Codel*, Hofmann*, Houghton*, Kuno*, Milani*,
  Mohanty*, Perez~Liebana*, Salakhutdinov*, Topin*, Veloso*, and
  Wang*]{gussminerlneurips2019}
William~H. Guss, Cayden Codel*, Katja Hofmann*, Brandon Houghton*, Noboru
  Kuno*, Stephanie Milani*, Sharada Mohanty*, Diego Perez~Liebana*, Ruslan
  Salakhutdinov*, Nicholay Topin*, Manuela Veloso*, and Phillip Wang*.
\newblock The {M}ine{RL} competition on sample efficient reinforcement learning
  using human priors.
\newblock In \emph{The 33rd Conference on Neural Information Processing Systems
  Competition Track}, 2019.

\bibitem[Guss* et~al.(2019)Guss*, Houghton*, Topin, Wang, Codel, Veloso, and
  Salakhutdinov]{gussminerlijcai2019}
William~H. Guss*, Brandon Houghton*, Nicholay Topin, Phillip Wang, Cayden
  Codel, Manuela Veloso, and Ruslan Salakhutdinov.
\newblock Mine{RL}: A large-scale dataset of {M}inecraft demonstrations.
\newblock In \emph{The 28th International Joint Conference on Artificial
  Intelligence}, 2019.

\bibitem[Guss et~al.(2021)Guss, Castro, Devlin, Houghton, Kuno, Loomis, Milani,
  Mohanty, Nakata, Salakhutdinov, et~al.]{guss2021minerl}
William~H Guss, Mario~Ynocente Castro, Sam Devlin, Brandon Houghton,
  Noboru~Sean Kuno, Crissman Loomis, Stephanie Milani, Sharada Mohanty, Keisuke
  Nakata, Ruslan Salakhutdinov, et~al.
\newblock The minerl 2020 competition on sample efficient reinforcement
  learning using human priors.
\newblock \emph{arXiv preprint arXiv:2101.11071}, 2021.

\bibitem[Harter et~al.(2020)Harter, Melnik, Kumar, Agarwal, Garg, and
  Ritter]{harter2020solving}
Augustin Harter, Andrew Melnik, Gaurav Kumar, Dhruv Agarwal, Animesh Garg, and
  Helge Ritter.
\newblock Solving physics puzzles by reasoning about paths.
\newblock In \emph{1st NeurIPS workshop on Interpretable Inductive Biases and
  Physically Structured Learning}, 2020.

\bibitem[He et~al.(2015)He, Zhang, Ren, and Sun]{DBLP:journals/corr/HeZRS15}
Kaiming He, Xiangyu Zhang, Shaoqing Ren, and Jian Sun.
\newblock Deep residual learning for image recognition.
\newblock \emph{CoRR}, abs/1512.03385, 2015.

\bibitem[Hessel et~al.(2018)Hessel, Modayil, Van~Hasselt, Schaul, Ostrovski,
  Dabney, Horgan, Piot, Azar, and Silver]{hessel2018rainbow}
Matteo Hessel, Joseph Modayil, Hado Van~Hasselt, Tom Schaul, Georg Ostrovski,
  Will Dabney, Dan Horgan, Bilal Piot, Mohammad Azar, and David Silver.
\newblock Rainbow: Combining improvements in deep reinforcement learning.
\newblock In \emph{Proceedings of the AAAI Conference on Artificial
  Intelligence}, 2018.

\bibitem[Hester et~al.(2018)Hester, Vecerik, Pietquin, Lanctot, Schaul, Piot,
  Horgan, Quan, Sendonaris, Osband, et~al.]{hester2018deep}
Todd Hester, Matej Vecerik, Olivier Pietquin, Marc Lanctot, Tom Schaul, Bilal
  Piot, Dan Horgan, John Quan, Andrew Sendonaris, Ian Osband, et~al.
\newblock Deep q-learning from demonstrations.
\newblock In \emph{Proceedings of the AAAI Conference on Artificial
  Intelligence}, 2018.

\bibitem[Hochreiter and Schmidhuber(1997)]{10.1162/neco.1997.9.8.1735}
Sepp Hochreiter and J\"{u}rgen Schmidhuber.
\newblock Long short-term memory.
\newblock \emph{Neural Comput.}, 9\penalty0 (8), 1997.

\bibitem[Houghton et~al.(2020)Houghton, Milani, Topin, Guss, Hofmann,
  Perez-Liebana, Veloso, and Salakhutdinov]{houghton2020guaranteeing}
Brandon Houghton, Stephanie Milani, Nicholay Topin, William Guss, Katja
  Hofmann, Diego Perez-Liebana, Manuela Veloso, and Ruslan Salakhutdinov.
\newblock Guaranteeing reproducibility in deep learning competitions.
\newblock In \emph{The 23rd Conference on Neural Information Processing
  Systems, Challenges in Machine Learning (CiML) Workshop}, 2020.

\bibitem[Juliani et~al.(2019)Juliani, Khalifa, Berges, Harper, Teng, Henry,
  Crespi, Togelius, and Lange]{juliani2019obstacle}
Arthur Juliani, Ahmed Khalifa, Vincent-Pierre Berges, Jonathan Harper, Ervin
  Teng, Hunter Henry, Adam Crespi, Julian Togelius, and Danny Lange.
\newblock Obstacle tower: A generalization challenge in vision, control, and
  planning.
\newblock \emph{arXiv preprint arXiv:1902.01378}, 2019.

\bibitem[Kang et~al.(2018)Kang, Jie, and Feng]{kang2018policy}
Bingyi Kang, Zequn Jie, and Jiashi Feng.
\newblock Policy optimization with demonstrations.
\newblock In \emph{International Conference on Machine Learning}, pages
  2469--2478. PMLR, 2018.

\bibitem[Khetarpal et~al.(2018)Khetarpal, Ahmed, Cianflone, Islam, and
  Pineau]{khetarpal2018re}
Khimya Khetarpal, Zafarali Ahmed, Andre Cianflone, Riashat Islam, and Joelle
  Pineau.
\newblock Re-evaluate: Reproducibility in evaluating reinforcement learning
  algorithms.
\newblock 2018.

\bibitem[Konen et~al.(2019)Konen, Korthals, Melnik, and
  Schilling]{konen2019biologically}
Kai Konen, Timo Korthals, Andrew Melnik, and Malte Schilling.
\newblock Biologically-inspired deep reinforcement learning of modular control
  for a six-legged robot.
\newblock In \emph{2019 IEEE International Conference on Robotics and
  Automation Workshop on Learning Legged Locomotion Workshop,(ICRA) 2019,
  Montreal, CA, May 20-25, 2019}, 2019.

\bibitem[K{\"o}nig et~al.(2018)K{\"o}nig, Melnik, Goeke, Gert, K{\"o}nig, and
  Kietzmann]{konig2018embodied}
Peter K{\"o}nig, Andrew Melnik, Caspar Goeke, Anna~L Gert, Sabine~U K{\"o}nig,
  and Tim~C Kietzmann.
\newblock Embodied cognition.
\newblock In \emph{2018 6th International Conference on Brain-Computer
  Interface (BCI)}, pages 1--4. IEEE, 2018.

\bibitem[Koppejan and Whiteson(2009)]{koppejan2009neuroevolutionary}
Rogier Koppejan and Shimon Whiteson.
\newblock Neuroevolutionary reinforcement learning for generalized helicopter
  control.
\newblock In \emph{Proceedings of the 11th Annual conference on Genetic and
  evolutionary computation}, pages 145--152, 2009.

\bibitem[Kumar et~al.(2020)Kumar, Zhou, Tucker, and
  Levine]{kumar2020conservative}
Aviral Kumar, Aurick Zhou, George Tucker, and Sergey Levine.
\newblock Conservative q-learning for offline reinforcement learning.
\newblock \emph{arXiv preprint arXiv:2006.04779}, 2020.

\bibitem[Lin(1992)]{lin1992self}
Long-Ji Lin.
\newblock Self-improving reactive agents based on reinforcement learning,
  planning and teaching.
\newblock \emph{Machine learning}, 8\penalty0 (3-4):\penalty0 293--321, 1992.

\bibitem[{Lloyd}(1982)]{lloyd1982}
S.~{Lloyd}.
\newblock Least squares quantization in pcm.
\newblock \emph{IEEE Transactions on Information Theory}, 28\penalty0
  (2):\penalty0 129--137, 1982.
\newblock \doi{10.1109/TIT.1982.1056489}.

\bibitem[Machado et~al.(2018)Machado, Bellemare, Talvitie, Veness, Hausknecht,
  and Bowling]{machado2018revisiting}
Marlos~C Machado, Marc~G Bellemare, Erik Talvitie, Joel Veness, Matthew
  Hausknecht, and Michael Bowling.
\newblock Revisiting the arcade learning environment: Evaluation protocols and
  open problems for general agents.
\newblock \emph{Journal of Artificial Intelligence Research}, 61:\penalty0
  523--562, 2018.

\bibitem[MacQueen et~al.(1967)]{macqueen1967some}
James MacQueen et~al.
\newblock Some methods for classification and analysis of multivariate
  observations.
\newblock 1967.

\bibitem[Malik et~al.(2021)Malik, Li, and Ravikumar]{malik2021generalizable}
Dhruv Malik, Yuanzhi Li, and Pradeep Ravikumar.
\newblock When is generalizable reinforcement learning tractable?
\newblock \emph{arXiv preprint arXiv:2101.00300}, 2021.

\bibitem[Marot et~al.(2020)Marot, Guyon, Donnot, Dulac-Arnold, Panciatici,
  Awad, O’Sullivan, Kelly, and Hampel-Arias]{marot2020l2rpn}
Antoine Marot, Isabelle Guyon, Benjamin Donnot, Gabriel Dulac-Arnold, Patrick
  Panciatici, Mariette Awad, Aidan O’Sullivan, Adrian Kelly, and Zigfried
  Hampel-Arias.
\newblock L2rpn: Learning to run a power network in a sustainable world
  neurips2020 challenge design.
\newblock 2020.

\bibitem[Meisheri et~al.(2019)Meisheri, Shelke, Verma, and
  Khadilkar]{meisheri2019accelerating}
Hardik Meisheri, Omkar Shelke, Richa Verma, and Harshad Khadilkar.
\newblock Accelerating training in pommerman with imitation and reinforcement
  learning.
\newblock \emph{arXiv preprint arXiv:1911.04947}, 2019.

\bibitem[Melnik et~al.(2018{\natexlab{a}})Melnik, Fleer, Schilling, and
  Ritter]{melnik2019modularization}
Andrew Melnik, Sascha Fleer, Malte Schilling, and Helge Ritter.
\newblock Modularization of end-to-end learning: Case study in arcade games.
\newblock In \emph{32nd Conference on Neural Information Processing Systems
  (NeurIPS 2018), Workshop on Causal Learning}, 2018{\natexlab{a}}.
\newblock URL \url{https://arxiv.org/pdf/1901.09895.pdf}.

\bibitem[Melnik et~al.(2018{\natexlab{b}})Melnik, Sch{\"u}ler, Rothkopf, and
  K{\"o}nig]{melnik2018world}
Andrew Melnik, Felix Sch{\"u}ler, Constantin~A Rothkopf, and Peter K{\"o}nig.
\newblock The world as an external memory: the price of saccades in a
  sensorimotor task.
\newblock \emph{Frontiers in behavioral neuroscience}, 12:\penalty0 253,
  2018{\natexlab{b}}.

\bibitem[Melnik et~al.(2019)Melnik, Bramlage, Voss, Rossetto, and
  Ritter]{melnik2019combining}
Andrew Melnik, Lennart Bramlage, Hendric Voss, Federico Rossetto, and Helge
  Ritter.
\newblock Combining causal modelling and deep reinforcement learning for
  autonomous agents in minecraft.
\newblock 2019.

\bibitem[Melnik et~al.(2021)Melnik, Harter, Limberg, and
  Ritter]{melnik2021critic}
Andrew Melnik, Augustin Harter, Christian Limberg, and Helge Ritter.
\newblock Critic-guided learning to segment rewarding objects in first-person
  views.
\newblock In \emph{NeurIPS 2020 Competition Track: The 2020 MineRL Competition
  on Sample Efficient Reinforcement Learning using Human Priors}, 2021.
\newblock URL \url{https://rebrand.ly/MineRLUNet}.

\bibitem[Milani et~al.(2020)Milani, Topin, Houghton, Guss, Mohanty, Nakata,
  Vinyals, and Kuno]{milani2020minerl}
Stephanie Milani, Nicholay Topin, Brandon Houghton, William~H Guss, Sharada~P
  Mohanty, Keisuke Nakata, Oriol Vinyals, and Noboru~Sean Kuno.
\newblock Retrospective analysis of the 2019 {MineRL} competition on sample
  efficient reinforcement learning.
\newblock \emph{Proceedings of the NeurIPS 2019 Competition and Demonstration
  Track}, 2020.

\bibitem[Mohanty et~al.(2020)Mohanty, Nygren, Laurent, Schneider, Scheller,
  Bhattacharya, Watson, Egli, Eichenberger, Baumberger,
  et~al.]{mohanty2020flatland}
Sharada Mohanty, Erik Nygren, Florian Laurent, Manuel Schneider, Christian
  Scheller, Nilabha Bhattacharya, Jeremy Watson, Adrian Egli, Christian
  Eichenberger, Christian Baumberger, et~al.
\newblock Flatland-rl: Multi-agent reinforcement learning on trains.
\newblock \emph{arXiv preprint arXiv:2012.05893}, 2020.

\bibitem[Nichol et~al.(2018)Nichol, Pfau, Hesse, Klimov, and
  Schulman]{nichol2018gotta}
Alex Nichol, Vicki Pfau, Christopher Hesse, Oleg Klimov, and John Schulman.
\newblock Gotta learn fast: A new benchmark for generalization in rl.
\newblock \emph{arXiv preprint arXiv:1804.03720}, 2018.

\bibitem[Oh et~al.(2016)Oh, Chockalingam, Singh, and Lee]{oh2016control}
Junhyuk Oh, Valliappa Chockalingam, Satinder Singh, and Honglak Lee.
\newblock Control of memory, active perception, and action in {M}inecraft.
\newblock \emph{arXiv preprint arXiv:1605.09128}, 2016.

\bibitem[Perez-Liebana et~al.(2019)Perez-Liebana, Hofmann, Mohanty, Kuno,
  Kramer, Devlin, Gaina, and Ionita]{perez2019multi}
Diego Perez-Liebana, Katja Hofmann, Sharada~Prasanna Mohanty, Noburu Kuno,
  Andre Kramer, Sam Devlin, Raluca~D Gaina, and Daniel Ionita.
\newblock The multi-agent reinforcement learning in {M}alm\"o ({MARL\"O})
  competition.
\newblock \emph{arXiv preprint arXiv:1901.08129}, 2019.

\bibitem[Pfeiffer et~al.(2018)Pfeiffer, Shukla, Turchetta, Cadena, Krause,
  Siegwart, and Nieto]{pfeiffer2018reinforced}
Mark Pfeiffer, Samarth Shukla, Matteo Turchetta, Cesar Cadena, Andreas Krause,
  Roland Siegwart, and Juan Nieto.
\newblock Reinforced imitation: Sample efficient deep reinforcement learning
  for mapless navigation by leveraging prior demonstrations.
\newblock \emph{IEEE Robotics and Automation Letters}, 3\penalty0 (4):\penalty0
  4423--4430, 2018.

\bibitem[Piot et~al.(2014)Piot, Geist, and Pietquin]{piot2014boosted}
Bilal Piot, Matthieu Geist, and Olivier Pietquin.
\newblock Boosted and reward-regularized classification for apprenticeship
  learning.
\newblock In \emph{Proceedings of the 2014 international conference on
  Autonomous agents and multi-agent systems}, pages 1249--1256, 2014.

\bibitem[Reddy et~al.(2020)Reddy, Dragan, and Levine]{DBLP:conf/iclr/ReddyDL20}
Siddharth Reddy, Anca~D. Dragan, and Sergey Levine.
\newblock {SQIL:} imitation learning via reinforcement learning with sparse
  rewards.
\newblock In \emph{8th International Conference on Learning Representations,
  Addis Ababa, Ethiopia, April 26-30}, 2020.

\bibitem[Rolnick et~al.(2018)Rolnick, Ahuja, Schwarz, Lillicrap, and
  Wayne]{rolnick2018experience}
David Rolnick, Arun Ahuja, Jonathan Schwarz, Timothy~P Lillicrap, and Greg
  Wayne.
\newblock Experience replay for continual learning.
\newblock \emph{arXiv preprint arXiv:1811.11682}, 2018.

\bibitem[Schaul et~al.(2015)Schaul, Quan, Antonoglou, and
  Silver]{schaul2015prioritized}
Tom Schaul, John Quan, Ioannis Antonoglou, and David Silver.
\newblock Prioritized experience replay.
\newblock In \emph{Proceedings of the International Conference on Learning
  Representations}, 2015.

\bibitem[Scheller et~al.(2020)Scheller, Schraner, and
  Vogel]{scheller2020sample}
Christian Scheller, Yanick Schraner, and Manfred Vogel.
\newblock Sample efficient reinforcement learning through learning from
  demonstrations in minecraft.
\newblock In \emph{NeurIPS 2019 Competition and Demonstration Track}, pages
  67--76, 2020.

\bibitem[Schilling and Melnik(2018)]{schilling2018approach}
Malte Schilling and Andrew Melnik.
\newblock An approach to hierarchical deep reinforcement learning for a
  decentralized walking control architecture.
\newblock In \emph{Biologically Inspired Cognitive Architectures Meeting},
  pages 272--282. Springer, 2018.

\bibitem[Shu et~al.(2017)Shu, Xiong, and Socher]{shu2017hierarchical}
Tianmin Shu, Caiming Xiong, and Richard Socher.
\newblock Hierarchical and interpretable skill acquisition in multi-task
  reinforcement learning.
\newblock \emph{arXiv preprint arXiv:1712.07294}, 2017.

\bibitem[Silver et~al.(2017)Silver, Schrittwieser, Simonyan, Antonoglou, Huang,
  Guez, Hubert, Baker, Lai, Bolton, et~al.]{silver2017mastering}
David Silver, Julian Schrittwieser, Karen Simonyan, Ioannis Antonoglou, Aja
  Huang, Arthur Guez, Thomas Hubert, Lucas Baker, Matthew Lai, Adrian Bolton,
  et~al.
\newblock Mastering the game of go without human knowledge.
\newblock \emph{Nature}, 550\penalty0 (7676):\penalty0 354--359, 2017.

\bibitem[Silver et~al.(2018)Silver, Hubert, Schrittwieser, Antonoglou, Lai,
  Guez, Lanctot, Sifre, Kumaran, Graepel, et~al.]{alphazero}
David Silver, Thomas Hubert, Julian Schrittwieser, Ioannis Antonoglou, Matthew
  Lai, Arthur Guez, Marc Lanctot, Laurent Sifre, Dharshan Kumaran, Thore
  Graepel, et~al.
\newblock A general reinforcement learning algorithm that masters chess, shogi,
  and go through self-play.
\newblock \emph{Science}, 362, 2018.

\bibitem[Simonyan et~al.(2013)Simonyan, Vedaldi, and
  Zisserman]{simonyan2013deep}
Karen Simonyan, Andrea Vedaldi, and Andrew Zisserman.
\newblock Deep inside convolutional networks: Visualising image classification
  models and saliency maps.
\newblock \emph{arXiv preprint arXiv:1312.6034}, 2013.

\bibitem[Stooke et~al.(2020)Stooke, Lee, Abbeel, and
  Laskin]{stooke2020decoupling}
Adam Stooke, Kimin Lee, Pieter Abbeel, and Michael Laskin.
\newblock Decoupling representation learning from reinforcement learning.
\newblock \emph{arXiv preprint arXiv:2009.08319}, 2020.

\bibitem[Tessler et~al.(2017)Tessler, Givony, Zahavy, Mankowitz, and
  Mannor]{tessler2017deep}
Chen Tessler, Shahar Givony, Tom Zahavy, Daniel~J Mankowitz, and Shie Mannor.
\newblock A deep hierarchical approach to lifelong learning in minecraft.
\newblock In \emph{The 31st AAAI Conference on Artificial Intelligence}, 2017.

\bibitem[Van~Hasselt et~al.(2016)Van~Hasselt, Guez, and Silver]{van2016deep}
Hado Van~Hasselt, Arthur Guez, and David Silver.
\newblock Deep reinforcement learning with double q-learning.
\newblock 2016.

\bibitem[Vinyals et~al.(2019{\natexlab{a}})Vinyals, Babuschkin, Czarnecki,
  Mathieu, Dudzik, Chung, Choi, Powell, Ewalds, Georgiev,
  et~al.]{starcraft2019}
Oriol Vinyals, Igor Babuschkin, Wojciech~M. Czarnecki, Michael Mathieu, Andrew
  Dudzik, Junyong Chung, David~H. Choi, Richard Powell, Timo Ewalds, Petko
  Georgiev, et~al.
\newblock Grandmaster level in {StarCraft II} using multi-agent reinforcement
  learning.
\newblock \emph{Nature}, 2019{\natexlab{a}}.

\bibitem[Vinyals et~al.(2019{\natexlab{b}})Vinyals, Babuschkin, Czarnecki,
  Mathieu, Dudzik, Chung, Choi, Powell, Ewalds, Georgiev,
  et~al.]{vinyals2019grandmaster}
Oriol Vinyals, Igor Babuschkin, Wojciech~M Czarnecki, Micha{\"e}l Mathieu,
  Andrew Dudzik, Junyoung Chung, David~H Choi, Richard Powell, Timo Ewalds,
  Petko Georgiev, et~al.
\newblock Grandmaster level in starcraft ii using multi-agent reinforcement
  learning.
\newblock \emph{Nature}, 575\penalty0 (7782):\penalty0 350--354,
  2019{\natexlab{b}}.

\bibitem[Wang et~al.(2016)Wang, Schaul, Hessel, Hasselt, Lanctot, and
  Freitas]{wang2016dueling}
Ziyu Wang, Tom Schaul, Matteo Hessel, Hado Hasselt, Marc Lanctot, and Nando
  Freitas.
\newblock Dueling network architectures for deep reinforcement learning.
\newblock In \emph{Proceedings of the International Conference on Machine
  Learning}, 2016.

\bibitem[Yarats et~al.(2021)Yarats, Kostrikov, and Fergus]{yarats2021image}
Denis Yarats, Ilya Kostrikov, and Rob Fergus.
\newblock Image augmentation is all you need: Regularizing deep reinforcement
  learning from pixels.
\newblock In \emph{International Conference on Learning Representations}, 2021.

\bibitem[Zhang et~al.(2018)Zhang, Ballas, and Pineau]{zhang2018dissection}
Amy Zhang, Nicolas Ballas, and Joelle Pineau.
\newblock A dissection of overfitting and generalization in continuous
  reinforcement learning.
\newblock \emph{arXiv preprint arXiv:1806.07937}, 2018.

\bibitem[Zhang et~al.(2019)Zhang, Dauphin, and
  Ma]{DBLP:journals/corr/abs-1901-09321}
Hongyi Zhang, Yann~N. Dauphin, and Tengyu Ma.
\newblock Fixup initialization: Residual learning without normalization.
\newblock \emph{CoRR}, abs/1901.09321, 2019.

\end{thebibliography}

\end{document}